\documentclass{article}

\usepackage[utf8]{inputenc}              
\usepackage[T1]{fontenc}                 
\usepackage{lmodern}                     
\usepackage{microtype}                   
\usepackage{amsmath, amssymb}            
\usepackage{graphicx}                    
\usepackage{booktabs}                    
\usepackage{hyperref}                    
\usepackage{xcolor}                      
\usepackage{geometry}                    
\usepackage[numbers]{natbib}             
\usepackage{authblk}                     
\usepackage{caption}                     
\usepackage{tcolorbox}                   
\usepackage{float}
\usepackage{etoolbox}
\makeatletter
\patchcmd{\thebibliography}{\clearpage}{}{}{}
\patchcmd{\thebibliography}{\cleardoublepage}{}{}{}
\makeatother
\tcbuselibrary{breakable}

\title{
  \vspace{-2cm} 
  \bfseries LLMs Can Teach Themselves to Better Predict the Future
}

\author[1]{Benjamin Turtel}
\author[1]{Danny Franklin}
\author[2]{Philipp Schoenegger}

\affil[1]{\href{https://www.lightningrod.ai/}{Lightning Rod Labs}}
\affil[2]{London School of Economics and Political Science}

\date{} 

\begin{document}

\maketitle

\begin{abstract}
We present an outcome-driven fine-tuning framework that enhances the forecasting capabilities of large language models (LLMs) without relying on human-curated reasoning samples. Our method leverages model self-play to generate pairs of diverse reasoning trajectories and probabilistic forecasts for a set of diverse questions that resolve after the models’ knowledge cutoff date. We then rank pairs of these reasoning traces by their distance to the actual outcomes before fine-tuning the model via Direct Preference Optimization (DPO). On a separate test set, our approach increases prediction accuracy of Phi-4 14B and DeepSeek-R1 14B by between 7--10\% over a base model and a DPO fine-tuned control model with randomized labels, bringing them on par with forecasting capabilities of much larger frontier models like GPT-4o.

\end{abstract}

\section{Introduction}
Large language models (LLMs) have demonstrated remarkable capabilities in a wide range of areas, often approaching or exceeding human performance. One area where human performance has not yet been surpassed is judgemental forecasting \cite{Karger2024}, where probabilistic forecasts are assigned to future events. Successful forecasts by top-performing human forecasters include substantial reasoning about facts of the world, various trends, and competing pieces of evidence \cite{Tetlock2016}, making it a great place to study model reasoning capabilities in a messy real-world environment.

Moreover, forecasting is a central task in decision-making across sectors as diverse as finance, policy, and law. It is central to inform resource allocation, manage risks, and plan organizational decisions. Modern LLMs have already been shown to conduct financial analysis \cite{Kim2024}, evaluate the impact of events on time series \cite{Wang2024}, and improve climate policy decision-making \cite{Cao2024}. This makes improving LLMs’ forecasting abilities potentially impactful and wide-ranging.

\begin{figure}[htbp]
  \centering
  \captionsetup{position=above} 
  \caption{Overview Flowchart}
  \includegraphics[width=0.8\linewidth]{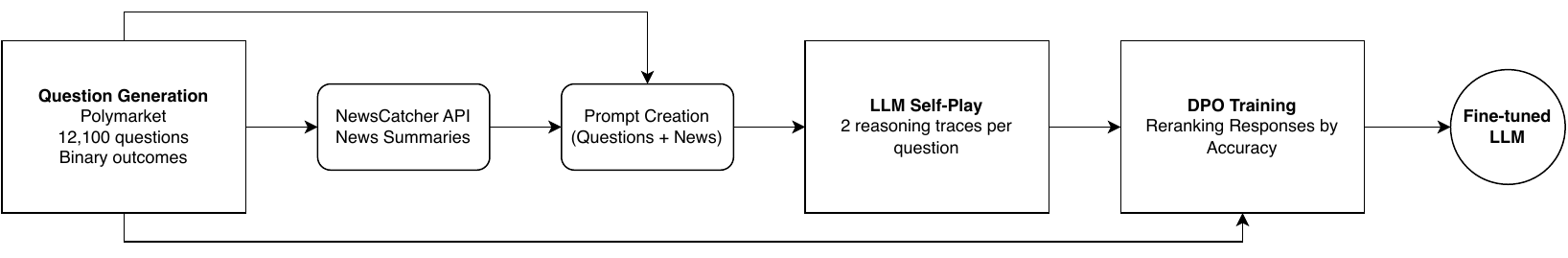}
  \captionsetup{position=below} 
  \caption*{\textbf{Notes:} This chart outlines the stages of our method.}
  \label{fig:flowchart}
\end{figure}

There has been some work explicitly looking to apply and boost the forecasting capabilities of LLMs. Such work has relied on aggregation \cite{Schoenegger2024}, retrieval of news as well as fine-tuning \cite{Halawi2024}, and ranked-based context retrieval \cite{Yan2023}, among other approaches \cite{Lyu2024}. While most of these systems improve performance to varying degrees, many share a common methodological limitation: They are frequently reliant on human-curated data such as up-to-date crowd forecasts or output curation, and often fail to have the models learn from resolved outcomes. Human outputs are slow and costly to procure, making it difficult to have models continually learn from them and improve.

In this paper, we propose a new approach to improving LLM forecasting performance that sidesteps the use of human inputs above and beyond real-world resolutions and enables the model to directly learn from actual outcomes and self-play. Self-play, where models compete against themselves, has previously been used in AlphaGo Zero to achieve superhuman performance \citep{Silver2017}, as well as more recent fine-tuning approaches like Self-Play fIne-tuNing (SPIN) \citep{Chen2024}. By allowing the model to produce reasonings and forecasts by itself on a large number of forecasting questions, this provides us with a large data set that we can then use for further training. As such, we do not rely on human-written forecasting rationales or predictions and instead only use model-generated reasoning, making this straightforwardly scalable.

\begin{figure}[htbp]
  \centering
  \captionsetup{position=above} 
  \caption{Accuracy Results for all Models}
  \includegraphics[width=0.8\linewidth]{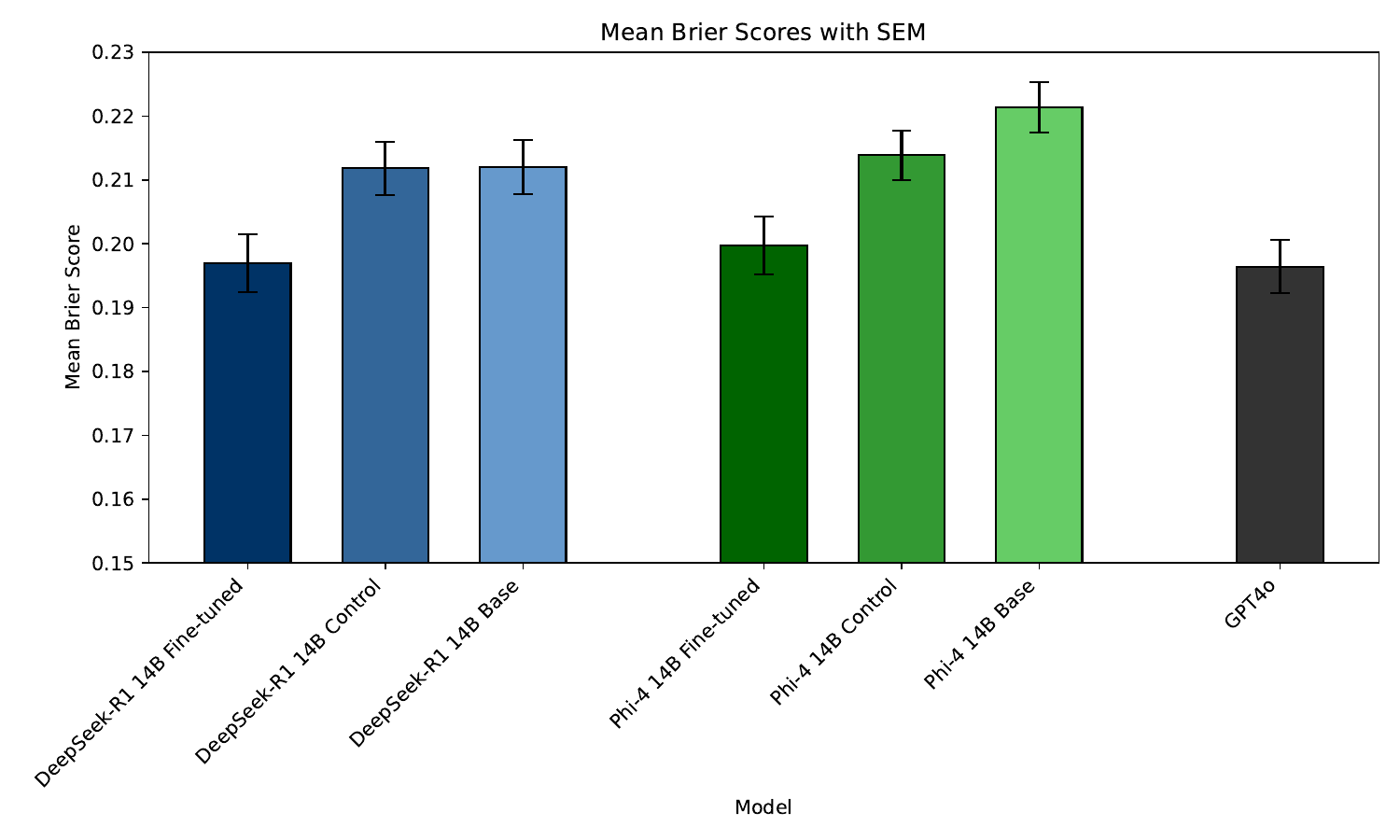}
  \captionsetup{position=below} 
  \caption*{\textbf{Notes:} The bar graph shows mean Brier scores with standard error of the mean (SEM) error bars. The y-axis starts at 0.15. Lower Brier scores indicate higher accuracy.}
  \label{fig:accuracy}
\end{figure}

Further, our approach uses Direct Preference Optimisation (DPO) \citep{Rafailov2024}, a reward-free method entirely bypassing the need for a separate reward model, to instead learn a reward signal from sets of ranked reasoning pairs \citep{Xu2024} drawn from the self-play outputs. This allows DPO to capitalize on relative rankings between forecasts, enabling the model to learn from the entire set of generated samples without the need for manual curation. Even when forecasts are individually suboptimal, DPO trains the model to discern subtle differences in quality and systematically correct biases through pairwise comparisons. By contrast, Supervised Fine-Tuning (SFT) relies on human-curated examples and treats selected forecasts as fully correct, which can lead to the discarding of potentially valuable information; DPO overcomes this limitation by learning from all samples, thereby enhancing the robustness and efficiency of the fine-tuning process.

Our work follows up on recent advances made by DeepSeek’s release of R1 \citep{Guo2025}, which demonstrated the power of reinforcement learning in deterministic contexts like mathematics and coding. We move the focus to real-world forecasting, which is inherently noisy and relies on calibrated predictions rather than simple binary correctness. This requires our models to learn from noisy probabilistic outcomes, which, if successful, promises widespread applicability.

To achieve this, we draw on a large dataset of resolved prediction market questions from Polymarket, where the model—restricted to a historical cutoff date—generates multiple reasoning traces and probabilistic forecasts through self-play. We then rank these pairs of rationales based on their proximity to the resolved outcome (for instance, ranking a 5\% prediction higher than a 10\% prediction if the event resolved to "No") before fine-tuning our model on them and testing the model on a separate test set. This ensures that the model does not simply learn whether a forecast predicted that an event would or would not occur, but instead enables it to draw directly from the full set of forecasts needed for a well-calibrated forecasting model (see Figure~\ref{fig:flowchart}).

Our results on a temporally held-out test set of questions resolving after December 25, 2024 show that for both of the models that we employed our method on, Phi-4 14B \citep{Abdin2024} and DeepSeek-R1 14B \citep{Guo2025}, we find accuracy improvements of between 7--10\% over the base versions of these models as well as the same models fine-tuned with randomized outcome labels as a control, see Figure~\ref{fig:accuracy}. Comparing not only to the base model but also to randomized-label fine-tuned controls allows us to more rigorously tease out the effect of outcome-based learning versus exposure to additional information. This shows that our method improves model forecasting performance, underscoring the potential of drawing on self-play reasoning data in improving probabilistic reasoning and prediction accuracy. Strikingly, our fine-tune of both models are also on par with the performance of the much larger GPT-4o \citep{Hurst2024}.

\section{Method}

Our approach consists of six main steps: 2.1) Collection and preprocessing of forecasting data, 2.2) News collection, 2.3) Synthetic training data generation through base model self-play, 2.4) Resolution-driven re-ranking, 2.5) Direct Preference Optimization (DPO) fine-tuning, and 2.6) Forecasting test-set questions.

For this pipeline, we used two models for self-play and for the final forecasting process: Phi-4 \citep{Abdin2024} and DeepSeek-R1-Distill-Qwen-14B \citep{Guo2025}. Both models are small (at 14B parameters) but have shown strong performance on general science and coding benchmarks, sometimes rivalling GPT-4o \citep{Rein2023,Hendrycks2021,Guo2025}. DeepSeek-R1-Distill-Qwen-14B is a distilled model derived from Gwen2.5-14B \citep{Yang2024} fine-tuned with the reasoning patterns from DeepSeek-R1 \citep{Guo2025}. Throughout this paper, we refer to these models as Phi-4 14B and DeepSeek-R1 14B respectively.

\subsection{Data}

We collected a total of 12,100 binary outcome forecasting questions from the prediction market Polymarket. We excluded all outcomes with ambiguous resolutions and partitioned the data as follows: our training set included 9,800 questions that all resolved between July 1 and December 15, 2024, and our test set included 2,300 questions that all resolved between December 25, 2024 and January 23, 2025. We also collected the final outcomes for all questions, recording as ‘0’ all outcomes that did not happen and as ‘1’ all outcomes that did happen. See Table~\ref{tab:example} for example questions.

\begin{table}[htbp]
  \centering
  \caption{Example Questions with Outcomes}
  \label{tab:example}
  \begin{tabular}{ll}
    \toprule
    \textbf{Question} & \textbf{Outcome} \\
    \midrule
    Will Sam Altman attend presidential inauguration? & 1 \\
    FTX doesn't start payouts in 2024? & 1 \\
    Will Modi win reelection? & 1 \\
    Will Sebastian Korda reach the quarterfinals of the Australian Open? & 0 \\
    Pershing Square IPO in 2024? & 0 \\
    \bottomrule
  \end{tabular}
  
  \vspace{1ex}
  \textbf{Notes:} Table~\ref{tab:example} shows a set of questions in the test set as well as their outcomes, with 0 indicating a negative resolution and 1 a positive resolution.
\end{table}

To evaluate the accuracy of our probabilistic forecasts in this paper, we calculate Brier scores. For each forecasting question with a predicted probability $p_i$ and an actual outcome $o_i \in \{0,1\}$, the Brier score is defined as

\[
BS = \frac{1}{N}\sum (p_i - o_i)^2,
\]

where $N$ is the total number of forecasting questions. A lower Brier score indicates higher forecasting accuracy.

\subsection{News Collection}
We collected news via the NewsCatcher API 14 days prior to question resolution. Our approach was drawn from \cite{Halawi2024} in that we generated search queries via GPT-4o and then integrated external news retrieval services like Newscatcher to aggregate and process the output. These news articles were then used as further input in Sections 2.3 and 2.6.

\subsection{Model Self-Play Data Generation}
We then instructed the base models to provide reasoning and a final probabilistic forecast for each question. For Phi-4 14B, we employed a scratchpad prompt \citep{Nye2021}, while we used a zero-shot prompt for DeepSeek-R1 14B as \texttt{<think>} tags are already present in the model output generation. The prompt included a summary of news from Section 2.2 along with the appropriate scratchpad or zero-shot prompt depending on the model. We ran all queries with a temperature of 1. In total, we generated a pair of reasoning traces for each question \citep{Munos2023}. We first generated a single reasoning and then re-ran this process up to four times to arrive at a second, non-identical forecast. If all subsequent predictions were identical, we removed the full set of forecasts. Overall, we obtained 18,854 reasoning traces for the 9,427 forecasting questions that had non-constant forecasts.

\subsection{Resolution-Driven Re-Ranking}
For each question, we paired up reasoning–outcome pairs and ranked them based on the proximity of the probabilistic forecast (ranging from 0\% to 100\%) to the ground truth (0 or 1). Formally, for each question with ground truth $o \in \{0,1\}$, let the probabilistic forecasts from two reasoning traces be denoted by $p_1$ and $p_2$ (with $p_i \in [0,1]$). We then define a ranking metric as

\[
r(p,o) = |p - o|,
\]

which measures the absolute difference between the forecast and the actual outcome. For example, if a pair consists of reasonings with 4\% and 8\% predictions respectively --- i.e. $p_1 = 0.04$ and $p_2 = 0.08$ --- with a ground truth of 0, then

\[
r(0.04,0) = 0.04 \quad \text{and} \quad r(0.08,0) = 0.08.
\]

Since $0.04 < 0.08$, the reasoning trace resulting in the 4\% prediction is ranked above that of the reasoning resulting in the 8\% forecast. Notably, the squared error metric of the Brier score naturally mitigates overconfidence by penalizing large deviations more heavily. Pairs that resulted in identical forecasts (i.e. $p_1 = p_2$) were removed prior to this stage. In total, we used the full set of 18,854 reasoning traces for the 9,427 forecasting questions for our re-ranking.

Moreover, to control for the possibility that information provided via the news aggregation at this step might influence the rankings, we also fine-tuned a second set of models via the same process, but with the ranking of labels randomised. These control models allow us to test whether the learning is attributable to the models learning from the higher-accuracy forecasting rationales.

\subsection{Direct Preference Optimization Fine-Tuning}
\begin{figure}[!htbp]
  \centering
  \captionsetup{position=above} 
  \caption{Per-Epoch Accuracy.}
  \includegraphics[width=0.8\linewidth]{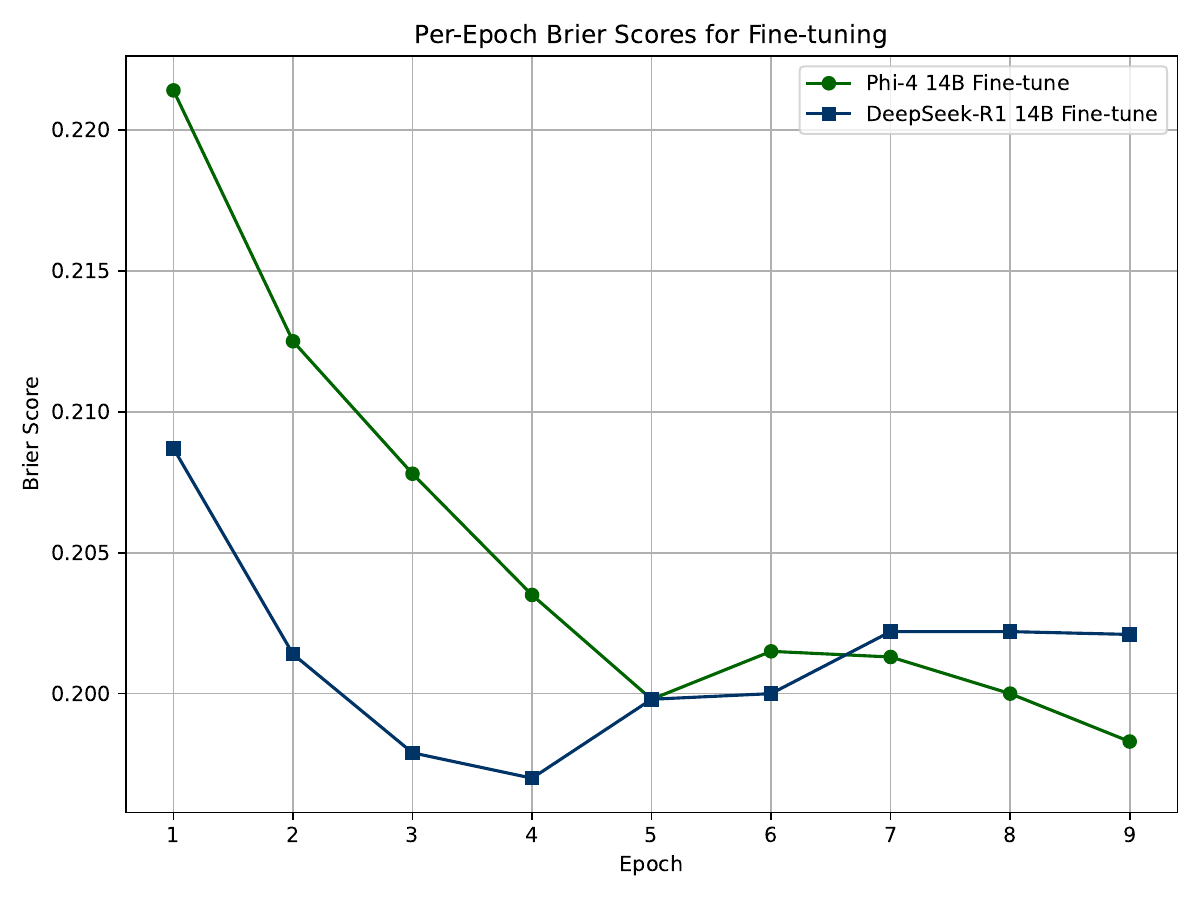}
  \captionsetup{position=below} 
  \caption*{\textbf{Notes:} This plot shows the per-epoch accuracy results for both Phi-4 14B and DeepSeek-R1 14B.}
  \label{fig:per_epoch}
\end{figure}

We then fine-tuned Phi-4 14B and DeepSeek-R1 14B using the preference pairs from Section 2.3. We use Direct Preference Optimization (DPO) to optimise model outputs against self-play derived and outcome-driven preferences without the need to train a separate reward model. The DPO loss was minimised using a LoRA adapter (rank=16, alpha=32, dropout=0.05, target\_modules="all-linear", no bias) on top of the base model, which was held in 4-bit quantisation, using a batch size of 2 (with 4 gradient accumulation steps) and gradient checkpointing enabled. Training leveraged the AdamW optimiser with a linear learning rate scheduler (5e-5 base rate), beta=0.1, and BF16 mixed precision. We used 8 H100 GPUs for training. For Phi-4 14B, we found a plateau at the fifth epoch, while this occurred at the fourth epoch for DeepSeek-R1 14B (see Figure~\ref{fig:per_epoch}).

\subsection{Forecasting Test Set Questions}
Finally, we test every model against a held-out test set of 2300 questions. Importantly, this test set begins 10 days after the final outcome in the training set, so our fine-tuned models have not been exposed to any news that might inform outcomes in the test set.

We do this with three versions of each model: the original base model, the fine-tuned model with correct outcomes for DPO ranking, and a control fine-tuned model with randomized outcomes for DPO ranking. This allows us to distinguish between learning that happened due to exposure to new information (for example, the news articles shared in prompts) versus learning by optimising for reasoning processes that lead to more accurate forecasts.

To generate our final forecasts, we used the following prompts shown in Figure~\ref{fig:prompts}, derived from Halawi et al. \cite{Halawi2024}. Our prompts drew on expert persona prompting \cite{XuExpertprompting2023}, based on structured analytic techniques \cite{Pherson2019} and Tetlock-style superforecasting \cite{Tetlock2016}, as well as more structured instructions, aiming to improve forecasting accuracy over a naïve assistant prompt.

\captionsetup{type=figure}
\captionof{figure}{Forecasting Prompts by Model}\label{fig:prompts}

\begin{tcolorbox}[breakable, colback=white, colframe=black, sharp corners, boxrule=1pt]

\textbf{Phi-4 14B:}

[Question, Question Background, Resolution Criteria, Today’s/Question Close Date, News Summaries]

\textbf{Instructions:}

1. Given the above question, rephrase and expand it to help you do better answering. Maintain all information in the original question.

{ Insert rephrased and expanded question. }

2. Using your knowledge of the world and topic, as well as the information provided, provide a few reasons why the answer might be no. Rate the strength of each reason.

{ Insert your thoughts }

3. Using your knowledge of the world and topic, as well as the information provided, provide a few reasons why the answer might be yes. Rate the strength of each reason.

{ Insert your thoughts }

4. Aggregate your considerations. Think like a superforecaster (e.g. Nate Silver).

{ Insert your aggregated considerations }

5. Output an initial probability (prediction) given steps 1--4.

{ Insert initial probability. }

6. Evaluate whether your calculated probability is excessively confident or not confident enough. Also, consider anything else that might affect the forecast that you did not before consider (e.g. base rate of the event).

{ Insert your thoughts }

7. Output your final prediction (a number between 0 and 1) with an asterisk at the beginning and end of the decimal.

{ Insert your answer }

\medskip

\textbf{DeepSeek R1 14B:}

You are an expert superforecaster, familiar with Structured Analytic Techniques as well as \textit{Superforecasting} by Philip Tetlock and related work. Predict the probability that the following question will be resolved as true/yes. You MUST give a probability estimate between 0 and 1 UNDER ALL CIRCUMSTANCES.

[Question, Question Background, Resolution Criteria, Today’s/Question Close Date, and News Summaries]

Output your final prediction (a number between 0 and 1) with an asterisk at the beginning and end of the decimal (Ex: *\textless probability\textgreater*).

{ Insert your answer }
\end{tcolorbox}

Both models were provided with the question, the question background, resolution criteria, the current date, the date when the forecasting question closes, and a summary of up to 10 news articles. We then collected forecasts for each model on the entire test set of 2300 questions. All models provided valid forecasts on all questions.

\section{Results}

\begin{figure}[htbp]
  \centering
  \captionsetup{position=above} 
  \caption{Ridge Plot of Forecasting Accuracy for each Model.}
  \includegraphics[width=0.8\linewidth]{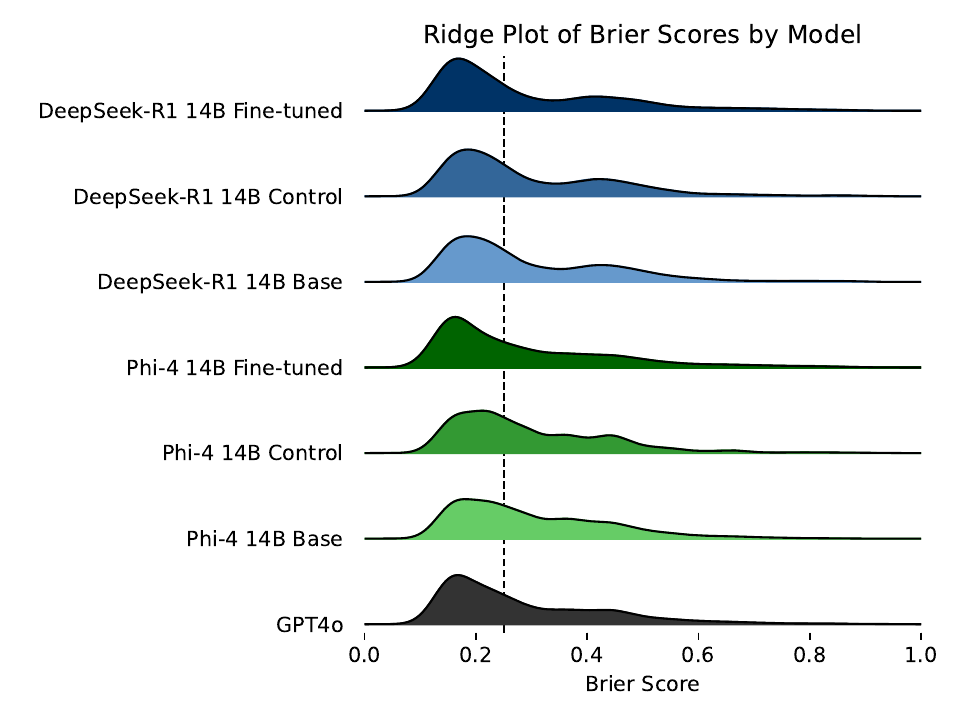}
  \captionsetup{position=below} 
  \caption*{\textbf{Notes:} The ridge plot displays the kernel density estimates of Brier scores for each model. Lower Brier scores indicate higher accuracy. The dotted vertical black line represents the ignorance benchmark of assigning 50\% to every question at a Brier score of 0.25.}
  \label{fig:ridge}
\end{figure}

For all results below, we call the fine-tuned model ‘Fine-Tune’, the base model ‘Base’, and the fine-tuned model with randomized labels the ‘Control’. We find substantial improvements in forecasting accuracy for both Phi-4 14B and DeepSeek-R1 14B fine-tunes, heavily outperforming the ignorance benchmark of a Brier score of 0.25 (arrived at by predicting 50\% on every question) and improving upon the base and control models (see Figure~\ref{fig:ridge}).

For Phi‑4 14B, the fine‑tuned model achieved a mean Brier score of 0.200 (SD = 0.218; 95\% CI [0.191, 0.209]), outperforming both the randomized‑label control model (M = 0.214, SD = 0.186; 95\% CI [0.206, 0.221]) and the base model (M = 0.221, SD = 0.189; 95\% CI [0.214, 0.229]). Similarly, DeepSeek‑R1 14B attained a mean Brier score of 0.197 (SD = 0.218; 95\% CI [0.188, 0.206]) after fine‑tuning, surpassing both its randomized‑label control (M = 0.212, SD = 0.202; 95\% CI [0.204, 0.220]) and base counterparts (M = 0.211, SD = 0.201; 95\% CI [0.204, 0.220]).\footnote{We hypothesise that one reason why the Phi-4 14B control model improves over the base, whereas there is no such effect for DeepSeek-R1 14B, is that it is likely to learn significantly more from the news articles—even with randomised labels—because it has a much earlier knowledge cut-off than DeepSeek-R1 14B.}

\begin{table}[htbp]
  \centering
  \caption{Results - Descriptive Statistics}
  \label{tab:descriptive}
  \begin{tabular}{lcccc}
    \toprule
    \textbf{Model} & \textbf{Mean Brier Score} & \textbf{SD} & \textbf{SEM} & \textbf{95\% CI} \\
    \midrule
    \multicolumn{5}{l}{\textbf{Phi-4 14B}} \\
    Fine-Tune & 0.200 & 0.218 & 0.005 & [0.191, 0.209] \\
    Control   & 0.214 & 0.186 & 0.004 & [0.206, 0.221] \\
    Base      & 0.221 & 0.189 & 0.004 & [0.214, 0.229] \\
    \midrule
    \multicolumn{5}{l}{\textbf{DeepSeek-R1 14B}} \\
    Fine-Tune & 0.197 & 0.218 & 0.005 & [0.188, 0.206] \\
    Control   & 0.212 & 0.202 & 0.004 & [0.204, 0.220] \\
    Base      & 0.212 & 0.201 & 0.004 & [0.204, 0.220] \\
    \midrule
    \multicolumn{5}{l}{\textbf{Frontier Benchmark}} \\
    GPT-4o    & 0.196 & 0.200 & 0.004 & [0.188, 0.205] \\
    \bottomrule
  \end{tabular}
  
  \vspace{1ex}
  \textbf{Notes:} Descriptive statistics for each model, including mean Brier scores, standard deviation, standard error of the mean, and 95\% confidence intervals. The sample size for all models is 2300 questions.
\end{table}

We conduct independent samples t-tests between the fine-tuned versions of the models and both the base and control models, as well as the frontier model benchmark set by GPT-4o. We find that for both Phi-4 14B and DeepSeek-R1 14B, the fine-tuned model is statistically significantly more accurate than both the base and control models at $p<0.05$. This also holds after adjusting the p-values for multiple comparisons via the Benjamini-Hochberg procedure \citep{Benjamini1995}. This suggests that our method is able to robustly and consistently improve forecasting performance, and that this performance increase is not due to the additional information that fine-tuning on the reasoning traces brings.

However, we fail to observe statistically significant differences between the fine-tuned models and the frontier model benchmark set by GPT-4o, $p>0.7$ for both after adjustment. The fact that GPT-4o does not outperform our small fine-tuning models shows that our method was effective in producing forecasting performance on par with much larger frontier models. Our usage of 4-bit quantization, which typically leads to small-to-medium performance reductions \citep{Zem2024, Huang2024}, further shows that our results are competitive even under these constraints.

\begin{table}[htbp]
  \centering
  \caption{Pairwise Comparisons with Adjusted p-values}
  \label{tab:pairwise}
  \begin{tabular}{l l c c}
    \toprule
    \textbf{Model 1} & \textbf{Model 2} & \textbf{p-value} & \textbf{Adj. p-value} \\
    \midrule
    DeepSeek-R1 14B Fine-tune & DeepSeek-R1 14B Base    & 0.015 & 0.027 \\
    DeepSeek-R1 14B Fine-tune & DeepSeek-R1 14B Control & 0.017 & 0.027 \\
    DeepSeek-R1 14B Fine-tune & GPT-4o                  & 0.931 & 0.931 \\
    Phi-4 14B Fine-tune       & Phi-4 14B Base          & 0.000 & 0.002 \\
    Phi-4 14B Fine-tune       & Phi-4 14B Control       & 0.018 & 0.027 \\
    Phi-4 14B Fine-tune       & GPT-4o                  & 0.589 & 0.706 \\
    \bottomrule
  \end{tabular}
  
  \vspace{1ex}
  \textbf{Notes:} The table shows p-values of independent samples t-tests. Adjustment of p-values is done via the Benjamini-Hochberg correction.
\end{table}

Comparing the distributions of accuracy scores across the questions for DeepSeek‑R1 14B, we find that the fine‑tuned model had a Brier score above 0.5 (very low accuracy) on 8.52\% of questions, slightly higher than the base (7.48\%) and control (7.61\%) models. However, it also had a Brier score below 0.05 (very high accuracy) on 32.78\% of questions, compared to only 23.22\% and 23.13\% for the base and control models. This indicates that while the fine‑tuned model occasionally makes slightly more highly inaccurate forecasts, it produces far more extremely accurate ones, more than compensating for the small uptick in large errors. We replicate this pattern at a similar magnitude for Phi‑4 14B, where the fine‑tuned model has 8.87\% of forecasts above 0.5 but 35.7\% below 0.05, compared to 7.26\% and 21\% for the base model and 6.43\% and 20.39\% for the control model, respectively.

\section{Conclusion}
Large language models can enhance their forecasting capabilities through self-play, generating reasoning traces that enable outcome-based fine-tuning without relying on human-curated data. By pairing these traces and ranking them by their proximity to actual outcomes, the models learn to refine their probabilistic forecasts, outperforming base models and matching the performance of larger frontier models.

\bibliography{Ref}

\begin{thebibliography}{26}
\providecommand{\natexlab}[1]{#1}
\providecommand{\url}[1]{\texttt{#1}}
\expandafter\ifx\csname urlstyle\endcsname\relax
  \providecommand{\doi}[1]{doi: #1}\else
  \providecommand{\doi}{doi: \begingroup \urlstyle{rm}\Url}\fi

\bibitem[Karger et~al.(2024)Karger, Bastani, Yueh-Han, Jacobs, Halawi, Zhang, and Tetlock]{Karger2024}
E.~Karger, H.~Bastani, C.~Yueh-Han, Z.~Jacobs, D.~Halawi, F.~Zhang, and P.~E. Tetlock.
\newblock Forecastbench: A dynamic benchmark of ai forecasting capabilities, 2024.
\newblock arXiv preprint arXiv:2409.19839.

\bibitem[Tetlock and Gardner(2016)]{Tetlock2016}
P.~E. Tetlock and D.~Gardner.
\newblock \emph{Superforecasting: The art and science of prediction}.
\newblock Random House, 2016.

\bibitem[Kim et~al.(2024)Kim, Muhn, and Nikolaev]{Kim2024}
A.~Kim, M.~Muhn, and V.~Nikolaev.
\newblock Financial statement analysis with large language models, 2024.
\newblock arXiv preprint arXiv:2407.17866.

\bibitem[Wang et~al.(2024)Wang, Feng, Qiu, Gu, and Zhao]{Wang2024}
X.~Wang, M.~Feng, J.~Qiu, J.~Gu, and J.~Zhao.
\newblock From news to forecast: Integrating event analysis in llm-based time series forecasting with reflection, 2024.
\newblock arXiv preprint arXiv:2409.17515.

\bibitem[Cao et~al.(2024)Cao, Zhuang, and He]{Cao2024}
C.~Cao, J.~Zhuang, and Q.~He.
\newblock Llm-assisted modeling and simulations for public sector decision-making: Bridging climate data and policy insights.
\newblock In \emph{AAAI-2024 Workshop on Public Sector LLMs: Algorithmic and Sociotechnical Design}, 2024.

\bibitem[Schoenegger et~al.(2024)Schoenegger, Tuminauskaite, Park, and Tetlock]{Schoenegger2024}
P.~Schoenegger, I.~Tuminauskaite, P.~S. Park, and P.~E. Tetlock.
\newblock Wisdom of the silicon crowd: Llm ensemble prediction capabilities rival human crowd accuracy, 2024.
\newblock arXiv preprint arXiv:2402.19379.

\bibitem[Halawi et~al.(2024)Halawi, Zhang, Yueh-Han, and Steinhardt]{Halawi2024}
D.~Halawi, F.~Zhang, C.~Yueh-Han, and J.~Steinhardt.
\newblock Approaching human-level forecasting with language models, 2024.
\newblock arXiv preprint arXiv:2402.18563.

\bibitem[Yan et~al.(2023)Yan, Seraj, He, Meng, and Sylvain]{Yan2023}
Q.~Yan, R.~Seraj, J.~He, L.~Meng, and T.~Sylvain.
\newblock Autocast++: Enhancing world event prediction with zero-shot ranking-based context retrieval, 2023.
\newblock arXiv preprint arXiv:2310.01880.

\bibitem[Lyu et~al.(2024)Lyu, Shridhar, Malaviya, Zhang, Elazar, Tandon, and Callison-Burch]{Lyu2024}
Q.~Lyu, K.~Shridhar, C.~Malaviya, L.~Zhang, Y.~Elazar, N.~Tandon, and C.~Callison-Burch.
\newblock Calibrating large language models with sample consistency, 2024.
\newblock arXiv preprint arXiv:2402.13904.

\bibitem[Silver et~al.(2017)Silver, Hubert, Schrittwieser, Antonoglou, Lai, Guez, and Hassabis]{Silver2017}
D.~Silver, T.~Hubert, J.~Schrittwieser, I.~Antonoglou, M.~Lai, A.~Guez, and D.~Hassabis.
\newblock Mastering chess and shogi by self-play with a general reinforcement learning algorithm, 2017.
\newblock arXiv preprint arXiv:1712.01815.

\bibitem[Chen et~al.(2024)Chen, Deng, Yuan, Ji, and Gu]{Chen2024}
Z.~Chen, Y.~Deng, H.~Yuan, K.~Ji, and Q.~Gu.
\newblock Self-play fine-tuning converts weak language models to strong language models, 2024.
\newblock arXiv preprint arXiv:2401.01335.

\bibitem[Rafailov et~al.(2024)Rafailov, Sharma, Mitchell, Manning, Ermon, and Finn]{Rafailov2024}
R.~Rafailov, A.~Sharma, E.~Mitchell, C.~D. Manning, S.~Ermon, and C.~Finn.
\newblock Direct preference optimization: Your language model is secretly a reward model.
\newblock In \emph{Advances in Neural Information Processing Systems}, volume~36, 2024.

\bibitem[Xu et~al.(2024)Xu, Fu, Gao, Ye, Liu, Mei, and Wu]{Xu2024}
S.~Xu, W.~Fu, J.~Gao, W.~Ye, W.~Liu, Z.~Mei, and Y.~Wu.
\newblock Is dpo superior to ppo for llm alignment? a comprehensive study, 2024.
\newblock arXiv preprint arXiv:2404.10719.

\bibitem[Guo et~al.(2025)Guo, Yang, Zhang, Song, Zhang, Xu, and He]{Guo2025}
D.~Guo, D.~Yang, H.~Zhang, J.~Song, R.~Zhang, R.~Xu, and Y.~He.
\newblock Deepseek-r1: Incentivizing reasoning capability in llms via reinforcement learning, 2025.
\newblock arXiv preprint arXiv:2501.12948.

\bibitem[Abdin et~al.(2024)Abdin, Aneja, Behl, Bubeck, Eldan, Gunasekar, and Zhang]{Abdin2024}
M.~Abdin, J.~Aneja, H.~Behl, S.~Bubeck, R.~Eldan, S.~Gunasekar, and Y.~Zhang.
\newblock Phi-4 technical report, 2024.
\newblock arXiv preprint arXiv:2412.08905.

\bibitem[Hurst et~al.(2024)Hurst, Lerer, Goucher, Perelman, Ramesh, Clark, and Kivlichan]{Hurst2024}
A.~Hurst, A.~Lerer, A.~P. Goucher, A.~Perelman, A.~Ramesh, A.~Clark, and I.~Kivlichan.
\newblock Gpt-4o system card, 2024.
\newblock arXiv preprint arXiv:2410.21276.

\bibitem[Rein et~al.(2023)Rein, Hou, Stickland, Petty, Pang, Dirani, and Bowman]{Rein2023}
D.~Rein, B.~L. Hou, A.~C. Stickland, J.~Petty, R.~Y. Pang, J.~Dirani, and S.~R. Bowman.
\newblock Gpqa: A graduate-level google-proof q\&a benchmark, 2023.
\newblock arXiv preprint arXiv:2311.12022.

\bibitem[Hendrycks et~al.(2021)Hendrycks, Burns, Kadavath, Arora, Basart, Tang, and Steinhardt]{Hendrycks2021}
D.~Hendrycks, C.~Burns, S.~Kadavath, A.~Arora, S.~Basart, E.~Tang, and J.~Steinhardt.
\newblock Measuring mathematical problem solving with the math dataset, 2021.
\newblock arXiv preprint arXiv:2103.03874.

\bibitem[Yang et~al.(2024)Yang, Yang, Zhang, Hui, Zheng, Yu, and Qiu]{Yang2024}
A.~Yang, B.~Yang, B.~Zhang, B.~Hui, B.~Zheng, B.~Yu, and Z.~Qiu.
\newblock Qwen2.5 technical report, 2024.
\newblock arXiv preprint arXiv:2412.15115.

\bibitem[Nye et~al.(2021)Nye, Andreassen, Gur-Ari, Michalewski, Austin, Bieber, and Odena]{Nye2021}
M.~Nye, A.~J. Andreassen, G.~Gur-Ari, H.~Michalewski, J.~Austin, D.~Bieber, and A.~Odena.
\newblock Show your work: Scratchpads for intermediate computation with language models, 2021.
\newblock arXiv preprint arXiv:2112.00114.

\bibitem[Munos et~al.(2023)Munos, Valko, Calandriello, Azar, Rowland, Guo, Tang, Geist, Mesnard, Fiegel, Michi, Selvi, Girgin, Momchev, Bachem, Mankowitz, Precup, and Piot]{Munos2023}
R.~Munos, M.~Valko, D.~Calandriello, M.~G. Azar, M.~Rowland, Z.~D. Guo, Y.~Tang, M.~Geist, T.~Mesnard, C.~Fiegel, A.~Michi, M.~Selvi, S.~Girgin, N.~Momchev, O.~Bachem, D.~J. Mankowitz, D.~Precup, and B.~Piot.
\newblock Nash learning from human feedback, 2023.
\newblock arXiv preprint arXiv:2312.00886.

\bibitem[Xu et~al.(2023)Xu, Yang, Lin, Wang, Zhou, Zhang, and Mao]{XuExpertprompting2023}
B.~Xu, A.~Yang, J.~Lin, Q.~Wang, C.~Zhou, Y.~Zhang, and Z.~Mao.
\newblock Expertprompting: Instructing large language models to be distinguished experts, 2023.
\newblock arXiv preprint arXiv:2305.14688.

\bibitem[Pherson and Heuer(2019)]{Pherson2019}
R.~H. Pherson and R.~J. Heuer.
\newblock \emph{Structured analytic techniques for intelligence analysis}.
\newblock Cq Press, 2019.

\bibitem[Benjamini and Hochberg(1995)]{Benjamini1995}
Y.~Benjamini and Y.~Hochberg.
\newblock Controlling the false discovery rate: a practical and powerful approach to multiple testing.
\newblock \emph{Journal of the Royal Statistical Society: Series B (Methodological)}, 57\penalty0 (1):\penalty0 289--300, 1995.

\bibitem[Zem(2024)]{Zem2024}
O.~Zem.
\newblock Exploring the impact of quantization on llm performance.
\newblock \url{https://medium.com/@olga.zem/exploring-the-impact-of-quantization-on-llm-performance-5698e16c5564}, January~3 2024.
\newblock Accessed: 2024-01-03.

\bibitem[Huang et~al.(2024)Huang, Zheng, Ma, Qin, Lv, Chen, and Magno]{Huang2024}
W.~Huang, X.~Zheng, X.~Ma, H.~Qin, C.~Lv, H.~Chen, and M.~Magno.
\newblock An empirical study of llama3 quantization: From llms to mllms.
\newblock \emph{Visual Intelligence}, 2\penalty0 (1):\penalty0 36, 2024.

\end{thebibliography}

\end{document}